\documentclass{article}

\usepackage{arxiv}

\usepackage[utf8]{inputenc} 
\usepackage[T1]{fontenc}    
\usepackage{hyperref}       
\usepackage{url}            
\usepackage{booktabs}       
\usepackage{amsfonts}       
\usepackage{nicefrac}       
\usepackage{microtype}      
\usepackage{lipsum}		
\usepackage{graphicx}
\usepackage{natbib}
\usepackage{doi}
\usepackage{graphics}
\usepackage{xspace}
\usepackage{amsmath}
\usepackage{amssymb}
\usepackage{multirow}
\usepackage{multicol}
\usepackage{tabularx}
\usepackage{subcaption}

\title{$\name{}{}$: A Graph Reinforcement Learning Approach to Optimize Atomic Structures on Rough Energy Landscapes}

\author{Vaibhav Bihani,\\
Department of Civil Engineering, \\Indian Institute of Technology Delhi,\\
Hauz Khas, New Delhi, India, 110016\\
\texttt{ce1180169@civil.iitd.ac.in}\\
\And
Sahil Manchanda,\\
Department of Computer Science and Engineering, \\Indian Institute of Technology Delhi,\\
Hauz Khas, New Delhi, India 110016\\
\texttt{csz188551@cse.iitd.ac.in}\\
\And
Srikanth Sastry,\\
  Theoretical Sciences Unit and School of Advanced Materials,\\ Jawaharlal Nehru Centre for Advanced Scientific Research,\\
  Rachenahalli Lake Road, Bengaluru, India 560064\\
  \texttt{sastry@jncasr.ac.in}
\And
Sayan Ranu$^*$,\\
Department of Computer Science and Engineering, \\Indian Institute of Technology Delhi,\\ Hauz Khas, New Delhi, India 110016\\
\texttt{sayanranu@cse.iitd.ac.in}\\
\And
N. M. Anoop Krishnan\thanks{Yardi School of Artificial Intelligence, IIT Delhi}\\
 Department of Civil Engineering, \\Indian Institute of Technology Delhi,\\ Hauz Khas, New Delhi, India 110016\\
\texttt{krishnan@iitd.ac.in}
}
\date{}

\renewcommand{\headeright}{}
\renewcommand{\undertitle}{}
\renewcommand{\shorttitle}{\name{}: A graph RL to optimize atomic structures}

\begin{document}

\newcommand{\cff}{\textsc{Cf}$^2$\xspace}
\newcommand{\cfg}{\textsc{CF-GnnExplainer}\xspace}
\newcommand{\gem}{\textsc{Gem}\xspace}
\newcommand{\gnns}{\textsc{Gnn}s}
\newcommand{\gnn}{\textsc{Gnn}{}}
\newcommand{\gat}{\textsc{Gat}{}}
\newcommand{\MLP}{\texttt{MLP}{}}
\newcommand{\name}{\textsc{StriderNET}{}}
\newcommand{\softmax}{\texttt{SoftMax}{}}
\newcommand{\CG}{\mathcal{G}\xspace}
\newcommand{\CQ}{\mathcal{Q}\xspace}
\newcommand{\CM}{\mathcal{M}\xspace}
\newcommand{\CT}{\mathcal{T}\xspace}
\newcommand{\CB}{\mathcal{B}\xspace}
\newcommand{\CP}{\mathcal{P}\xspace}
\newcommand{\CV}{\mathcal{V}\xspace}
\newcommand{\CE}{\mathcal{E}\xspace}
\newcommand{\CC}{\mathcal{C}\xspace}
\newcommand{\CX}{\boldsymbol{\mathcal{X}\xspace}}
\newcommand{\CW}{\mathcal{W}\xspace}
\newcommand{\CZ}{\mathbf{Z}\xspace}
\newcommand{\CS}{\mathcal{S}\xspace}
\newcommand{\CF}{\mathcal{F}\xspace}
\newcommand{\CA}{\boldsymbol{\mathcal{A}}\xspace}
\newcommand{\cW}{\mathbf{W}\xspace}
\newcommand{\cS}{\mathbf{S}\xspace}
\newcommand{\CL}{\mathcal{L}\xspace}
\newcommand{\cx}{\mathbf{x}\xspace}
\newcommand{\ca}{\mathbf{a}\xspace}
\newcommand{\cy}{\mathbf{y}\xspace}
\newcommand{\ch}{\mathbf{h}\xspace}
\newcommand{\ce}{\mathbf{e}\xspace}
\newcommand{\cc}{\mathcal{c}\xspace}
\newcommand{\cz}{\mathbf{z}\xspace}
\newcommand{\bg}{\mathcal{B}\xspace}
\newcommand{\lhs}{\textsc{Lhs}\xspace}
\newcommand{\rhs}{\textsc{Rhs}\xspace}
\newcommand{\lb}{\textsc{Lb}\xspace}
\newcommand{\ub}{\textsc{Ub}\xspace}
\newcommand{\relu}{\text{ReLU}}
\newcommand{\gnode}{\textsc{Gnode}\xspace}
\newcommand{\node}{\textsc{Node}\xspace}
\newcommand{\lgn}{\textsc{Lgn}\xspace}
\newcommand{\hgn}{\textsc{Hgn}\xspace}
\newcommand{\mcgnode}{\textsc{MCGnode}\xspace}
\newcommand{\cgnode}{\textsc{CGnode}\xspace}
\newcommand{\cdgnode}{\textsc{CDGnode}\xspace}
\newcommand{\thatsymbol}{\fontencoding{T1}\selectfont \TH}
\newcommand{\expectation}{\mathbb{E}}
\newcommand{\flow}{\Gamma}
\newcommand{\diameter}{\mathcal{D}}
\newcommand{\pathset}{\mathbb{P}}
\newcommand{\Out}{\textbf{Out}}
\newcommand{\In}{\textbf{In}}
\newcommand{\vsa}{\vspace*{-0.2cm}}
\newcommand{\vsb}{\vspace*{-0.2cm}}
\newcommand{\vsc}{\vspace*{-0.4cm}}

\newcommand{\issue}[1]{\textcolor{ceruleanblue}{#1}}
\urlstyle{rm}

\newtheorem{thm}{\textbf{Theorem}}
\newtheorem{asm}{\textbf{Property}}
\newtheorem{defn}{\textbf{Definition}}
\newtheorem{lem}{\textbf{Lemma}}
\newtheorem{prop}{\textbf{Proposition}}
\newtheorem{cor}{\textbf{Corollary}}
\newtheorem{conc}{\textbf{Conclusion}}
\newtheorem{obs}{\textbf{Observation}}
\newtheorem{assump}{\textbf{Assumption}}
\newtheorem{example}{\textbf{Example}}
\newtheorem{prob}{\textbf{Problem}}
\newtheorem{claim}{\textbf{Claim}}
\newtheorem{advantage}{\textbf{Advantage}}
\newtheorem{facts}{\textbf{Fact}}
\newtheorem{discussion}{Discussion}
\newtheorem{theorem1}{Theorem}

\newcommand{\rev}[1]{\textcolor{blue}{#1}}

\maketitle
\begin{abstract}
  Optimization of atomic structures presents a challenging problem, due to their highly rough and non-convex energy landscape, with wide applications in the fields of drug design, materials discovery, and mechanics. Here, we present a graph reinforcement learning approach, $\name{}{}$, that learns a policy to displace the atoms towards low energy configurations. We evaluate the performance of $\name{}{}$ on three complex atomic systems, namely, binary Lennard-Jones particles, calcium silicate hydrates gel, and disordered silicon. We show that $\name{}{}$ outperforms all classical optimization algorithms and enables the discovery of a lower energy minimum. In addition, $\name{}{}$ exhibits a higher rate of reaching minima with energies, as confirmed by the average over multiple realizations. Finally, we show that $\name{}{}$ exhibits inductivity to unseen system sizes that are an order of magnitude different from the training system.
\end{abstract}


\section{Introduction and Related Work}

\begin{figure}[t]
    \centering
    \includegraphics[width=0.8\textwidth]{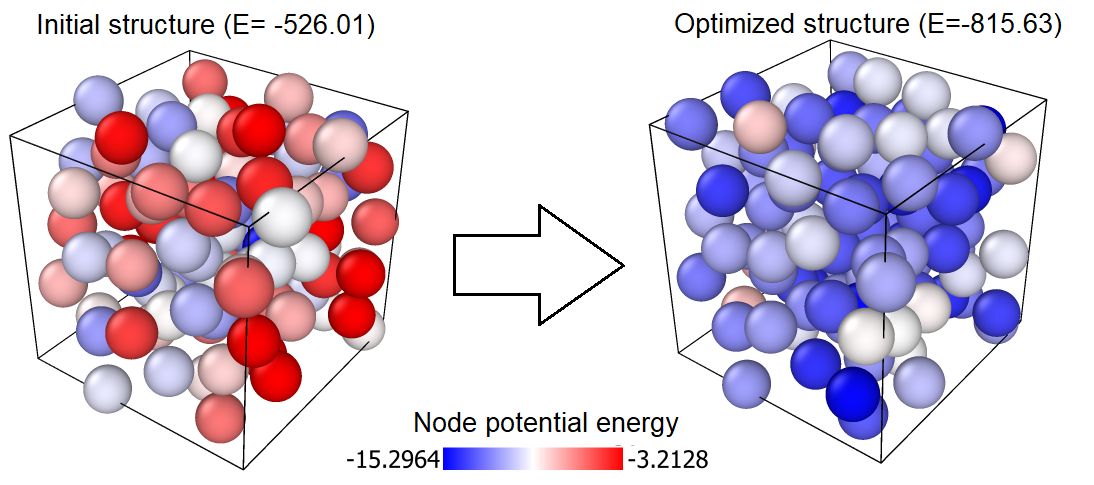}
    \caption{Optimization of $100$ atoms LJ system (Colorbar shows node potential energy).}
    \label{fig:LJ_optim_Node_pe}
\end{figure}

Optimization of functions exhibiting non-convex landscapes is a ubiquitous problem in several fields, such as the design of mechanical structures~\cite{mistakidis2013nonconvex}, robotics and motion planning~\cite{alonso2018cooperative,schwager2011unifying}, materials~\cite{le2016discovery}, and biological systems~\cite{yang2019machine}, such as proteins. Specifically, materials discovery relies on finding stable structures of atomic systems, such as new battery materials, novel drugs, or ultralight super-hard materials, through efficient optimization~\cite{xiang1995combinatorial}. These materials predicted through optimization are then verified and validated through experiments and tests for industrial applications. However, even for a given material having a few hundred atoms, a large number of possible structures can be obtained by allowing various configurational arrangements of the atoms. For instance, Fig.~\ref{fig:LJ_optim_Node_pe} shows the structure of a 100-atom Lennard-Jones system (detailed later), where the potential energy and positions of the atoms before and after optimization are shown. Extrapolation of previous work~\cite{tsai1993use} on simple atomic clusters suggests that a system containing 147 atoms can have as many as $10^{60}-10^{259}$ minima. These possible configurations of the atomic network, represented by local minima in the energy landscape separated by high energy barriers, make the optimization problem extremely challenging~\cite{wales2003energy}.
\looseness=-1

Several classical approaches have been proposed for optimization of atomic structures. These include \textit{fast inertial relaxation engine (FIRE)}~\cite{bitzek2006structural}, gradient-based approaches~\cite{stillinger1986local,leach2001molecular}, perturbation-based approaches~\cite{wales1997global}, and learned optimizers~\cite{merchant2021learn2hop}. However, most of these approaches present several drawbacks, namely, \textit{(i)} a significant number of iterations, \textit{(ii)} carefully hand-crafted update rules that are sensitive to parameters, \textit{(iii)} inability to scale to larger system sizes, \textit{(iv)} representation of atomic structures, and, most importantly, \textit{(v)} the inability to overcome high-energy barriers~\cite{wales2003energy}.
\looseness=-1

An alternative approach is to allow the system \textit{learn} policies that discover better minimum energy structures through \textit{reinforcement learning }(RL)~\cite{christiansen2020gaussian,simm2020reinforcement,rumelhart1986learning,meldgaard2020structure}. Most studies using RL for materials have focussed on small atomic clusters or simple molecules having a limited number of atoms. For extending the work to realistic structures, the first challenge is to develop a scalable representation of atomic structures. To this extent, graph neural networks (\gnn{s}) is an excellent choice---thanks to their ability to capture the local topology, while being inductive to unseen system sizes. \gnn{s} have been used extensively for modeling atomic and physical structures~\cite{batzner20223,bhattoo2023learning,thangamuthuunravelling,bhattoolearning,battaglia2018relational,bishnoi2022enhancing}. 
\looseness=-1

Here, we propose a framework combining \gnn{s} and RL, namely \textit{\name}\footnote{In our approach, RL trains the policy network to progressively take small \textit{strides} towards optimizing the graph representation of the atomic structure.}, that allows optimization of atomic structures exhibiting a rough energy landscape. Specifically, we show that combining a graph representation of atomic structures with a \textit{policy-gradient} approach outperforms the standard optimization algorithms. The main contributions of the present work are as follows.
\vspace{-0.05in}
\begin{itemize}
    \item \textbf{\name:} A graph reinforcement learning framework (Section~\ref{sec:Methodology}) that outperforms state-of-the-art optimizers on atomic structures (Section~\ref{sec:comparison}).
    \item \textbf{Graph matters:} The neighborhood information of atomic structure as captured by the graph architecture enables efficient optimization (Section~\ref{sec:graph_arch}). More importantly, a graph-based optimization framework for atomistic configurations has hitherto been unexplored, and this work initiates a new direction.
    \item \textbf{Model adaptation:} Adaptation of the model to a specific atomic structure allows the discovery of low energy states (Section~\ref{sec:adaptation}).
    \item \textbf{Inductivity:} The graph architecture allows the adaptation of a trained model to \textit{unseen} system sizes in an \textit{inductive} fashion (Section~\ref{sec:inductivity}).
\end{itemize}

\vspace{-0.10in}
\section{Preliminaries and Problem Formulation}
\label{sec:formulation}
\vspace{-0.05in}

This section introduces the preliminary concepts associated with the atomic structure optimization problem. 

The configuration $\Omega_c(\mathbf{x_1},\mathbf{x_2},...\mathbf{x}_N)$ of an atomic system is given by the positions of all the atoms in the system $(\mathbf{x_1}, \mathbf{x_2},\ldots,\mathbf{x}_N)$ and their types $\omega_i$. Each $\mathbf{x_i}$ represents the position of the $i^{th}$ atom in a $d$-dimensional space, where $d$ is typically $2$ or $3$. The potential energy $U$ of an $N$-atom structure is a function of $\Omega_c$. Specifically, the energy of a system can be written as the summation of one-body $U(r_i)$, two-body $U(r_i,r_j)$, three-body $U(r_i,r_j,r_k)$, up to $N$-body interaction terms as:

\begin{equation}
U = \sum_{i=1}^N U(r_i) + \sum_{\substack{i,j=1;\\i\neq j}}^{N} U(r_i,r_j) + \sum_{\substack{i,j,k=1;\\i\neq j \neq k}}^{N} U(r_i,r_j,r_k) + \cdots
\label{Eq:energy}
\end{equation}
However, the exact computation of this energy is highly challenging and involves expensive quantum mechanical computations~\cite{cohen2012challenges}. Alternatively, empirical potential functions~\cite{torrens2012interatomic} can approximately capture this interaction while maintaining the minima associated with these structures. These potentials are developed relying only on two-, three- or four-body interactions and ignoring higher-order terms for computational efficiency. In this work, we rely on well-validated empirical potentials to compute the energy of the different atomic structures. Accordingly, the atomic structure optimization can now be posed as a problem of identifying the configuration of $N$-atoms in terms of their position vectors, such that the system's total energy is minimum. 
\looseness=-1

The major challenge in such optimization is the rough landscape featuring an enormous number of stable structures (local minima) and a large number of degrees of freedom associated with an atomic structure ($Nd$ for an $N$-atom structure in $d$ dimensional space; typically $d=2$ or $3$). While characterizing the number of minima in the energy landscape of an actual material is challenging, several studies have been focuses on simple model systems. One of the classical systems extensively characterized includes the Lennard-Jones (LJ) system, which can be used to model noble gases~\cite{tsai1993use,wales1997global,malek2000dynamics,doye1999double}. The energy of a system of $N$-atoms interacting through the LJ potential is given by:
\begin{equation}
    U=\lambda \sum_{i=1}^{N-1}\sum_{\substack{j=2;\\j>i}}^N \left[\left(\frac{\beta}{|x_{ij}|}\right)^{12}-\left(\frac{\beta}{|x_{ij}|}\right)^6\right]
\label{eq:LJ}
\end{equation}
where $|x_{ij}|=|\mathbf{x_i}-\mathbf{x_j}|$ is the distance is between the atoms $i$ and $j$, and $\lambda$ and $\beta$ are constants depending on the atom types.
By extrapolating the studies on small LJ structures, the scaling of minima with the number of atoms $N$ can be obtained as $e^{(k_1+k_2N)}$ or $e^{(k_1+k_2N+k_3N^2)}$, where  $k_1, k_2$ and $k_3$ are constants obtained by fitting~\cite{wales1997global}. Thus, it becomes incredibly challenging for a system with thousands of atoms to get the global minima or even local minima with extremely low energy compared to the global minima.

Traditional approaches for optimizing atomic structures exploit the gradient of the energy $U$ with the positions to find stable structures near the starting configuration leading to local minima. Some of these approaches include steepest descent~\cite{stillinger1986local}, conjugate gradient, and Newton-Raphson~\cite{leach2001molecular}. Alternatively, FIRE relies on a momentum-based approach and has been shown to outperform purely gradient-based methods~\cite{bitzek2006structural}. These approaches aim to find the most stable atomic structure, starting from an arbitrary configuration. Thus, once trapped in a local minimum, these approaches cannot escape the minima to move toward more stable structures. Further, these approaches do not learn any new heuristics based on the trajectory they followed. Thus, there is no possibility of ``adapting" these algorithms to obtain more stable structures closer to the global minimum.
 To address these challenges, we propose a framework that exploits the atomic structure and energy relationship to discover stable configurations. 
\looseness=-1

\textbf{Problem: (Discovering stable structures)} \textit{Let $\Omega_c(\mathbf{x_1},\mathbf{x_2},...\mathbf{x_N})$ be a configuration of an $N$-atom system with energy $U^{\Omega_c}$ sampled from the energy landscape $\mathbb{U}^{Nd}$ of the system. Starting from $\Omega_c$, our goal is to obtain the configuration $\Omega_{min}$ exhibiting the minimum energy $U^{\Omega_{min}}$ by displacing the atoms. To this end, we aim to learn a policy $\pi$ that displaces the atom so that the system moves toward lower energy configurations while allowing it to overcome local energy barriers.}

In addition to the ability to find low-energy configurations, we also desire $\pi$ to satisfy the following properties:
 \begin{itemize}
     \item {\bf Permutation Invariance:} Policy $\pi$ is permutation invariant if $\pi(\Omega_c(\mathbf{x_1},\ldots,\mathbf{x_N}))=\pi(P(\Omega_c(\mathbf{x_1},\ldots,\mathbf{x_N})))$, where $P(\cdot)$ is a permutation over the constituent atoms. An atomistic configuration is a set of positions. Sets are permutation invariant by definition. Hence, if the policy is not permutation invariant, it will generate multiple representations for the same set (configuration) depending on the index ordering of atoms. This hampers generalizability to unseen configurations.
     \item {\bf Inductivity:} Policy $\pi$ is inductive if the number of parameters in the model is independent of $N$, i.e., the number of atoms in the system. If the policy is not inductive, it will be restricted to inference \textit{only} on atoms of size $N$, which limits generalizability to configurations of unseen sizes. As we will see later, the proposed methodology adopts a 2-phased learning procedure. First, we learn policy $\pi$ on atomic configurations of a given size. Now, given an unseen configuration of unseen size, we adapt the learned parameters for the input configuration. The ability to fine-tune learn parameters and optimize on any unseen configuration is feasible only due to the inductive nature of \name.
 \end{itemize}
 \begin{figure*}[t]
 \vspace{-0.10in}
    \centering
    \includegraphics[width=6.7in]{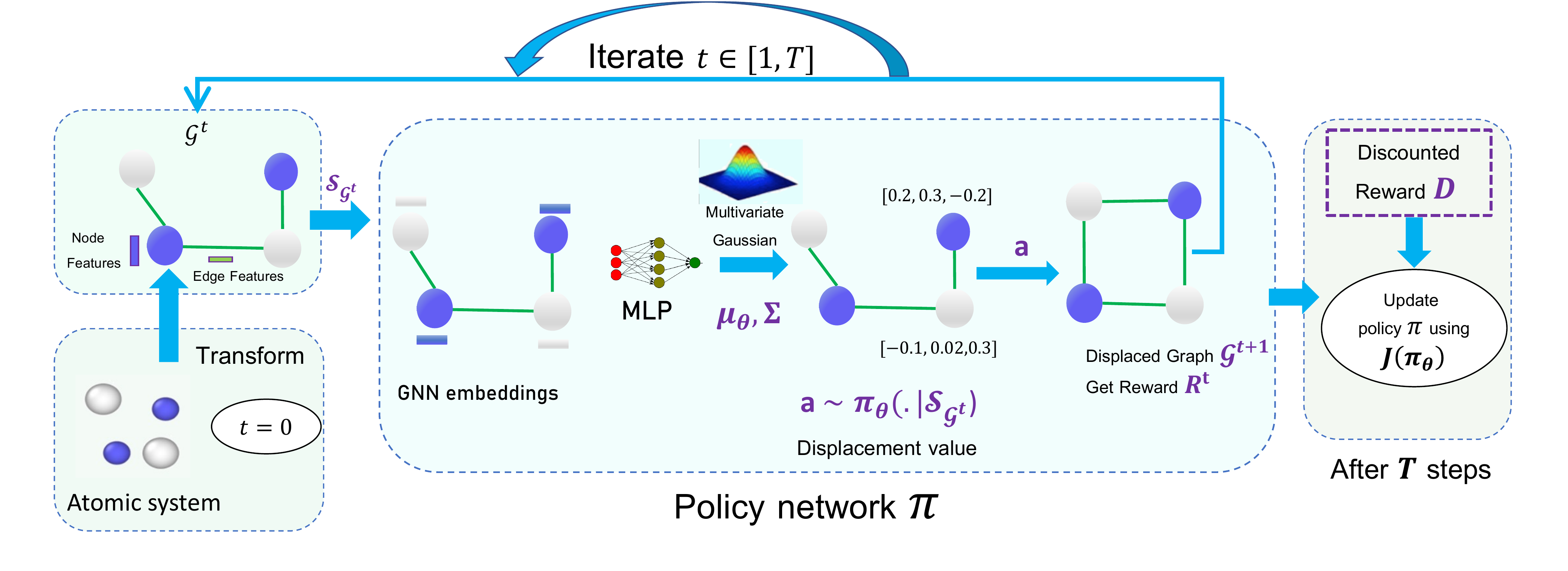}
    \caption{\name{}{} architecture.}
    \label{fig:arch}
    \vspace{-0.20in}
\end{figure*}
\section{\name{}: Proposed Methodology}
\label{sec:Methodology}
Fig.~\ref{fig:arch} describes the architecture of \name{}. To achieve the above-outlined objectives of permutation invariance and inductivity, we represent an atomistic configuration as a graph (more details in Section~\ref{sec:graph}). Subsequently, we develop a message-passing \gnn{} to embed graphs into a feature space. The message-passing architecture of the \gnn{} ensures both permutation invariance and inductivity. The graph, in turn, predicts the displacements of each of the atoms based on which the rewards are computed. Finally, the policy $\pi$ is learned by maximizing the discounted rewards. Note that we learn the parameters of $\pi$ using a set of training graphs exhibiting diverse energies that are sampled from the energy landscape $\mathbb{E}^{Nd}$ of an atomic system with $N$-atoms in $d$ dimensions. Thus, the initial structure, although arbitrary and possibly unstable, is realistic and physically feasible. 
Then given a new structure, we adapt the parameters of our learned policy network $\pi$ to the new structure while optimizing the new graph structure. All notations used in the present work are given in Tab.~\ref{tab:notation} in App.~\ref{sec:notation}.
 Before we define the parametrization of our policy, we first discuss how our atomic system is transformed into a graph. 
\vspace{-0.10in}
\subsection{Transforming atomic system to graph}
\label{sec:graph}
\vspace{-0.05in}
The total energy $U$ of an atomic system is closely related to the local neighborhood of an atom. In order to leverage this neighborhood information, we transform the atomic structure into a graph, where the nodes and edges of the graph represent the atoms and the chemical bonds between the atoms, respectively.
Thus, an atomic system is represented by a graph $\mathcal{G}=(\mathcal{V},\mathcal{E})$ where the nodes  $v 
\in \mathcal{V}$ denotes the atoms  and $e_{vu} \in \mathcal{E}$ represents edges corresponding to the interactions between atoms $v$ and $u$. Note that the edges can be dynamic in nature; new edges may form, or existing ones may break depending on the configuration $\Omega_i$.
Thus, the edges are defined for each $\Omega_c$ as a function of the distance between two nodes as $\mathcal{E}=\left\{e_{uv}=\left(u, v\right) \mid d\left(u, v\right) \leq \delta\right\}$ where $d\left(u, v\right)$ is a distance function over node positions and $\delta$ is a distance threshold. This threshold can be selected based on the first neighbor cutoff of the atomic structures as obtained from the pair-distribution function or based on the cutoff of the empirical potential. The cutoff thus defines the neighborhood of a node $v$ given by $\mathcal{N}_v=\{u| (u,v) \in \mathcal{E}\}$. 
\vspace{-0.10in}
\subsection{Learning policy $ \pi $ as Markov decision process}
\vspace{-0.05in}
Given an atomic structure represented as a graph $\mathcal{G}$ with the potential energy $U_\mathcal{G}$, our goal is to update the positions of the nodes $v \in \mathcal{V}$ for $t$ steps, such that the graph structure obtained after these updates $\mathcal{G}^t = (\mathcal{V},\mathcal{E}^t)$, has a lower potential energy $U_{\mathcal{G}^t}$. 
We model this task of iteratively updating the node positions as a \textit{Markov decision process} (MDP). Specifically, the \textit{state} is a function of the graph with its nodes and edges. The \textit{action} corresponds to displacing each of the nodes (atoms) in all $d$ directions as determined by policy $\pi$. The \textit{reward} is a function of  the change in potential energy obtained following the action(s) taken. In our case, we aim to decrease the potential energy of our given structure.  We next formalize each of these notions for our MDP formulation.\\
\textbf{State:} We denote the state of a graph $\mathcal{G}$ at step $t$ as a matrix $S_{\mathcal{G}^t}$, where the $i^{th}$ row in the matrix corresponds to the input node representation for the $i^{th}$ node. Intuitively, the state should contain information that would help our model make a decision regarding the magnitude and direction of each node's displacement. In this context, we note that the overall potential energy of the system is a function of the potential energy of individual atoms\footnote{We use the terms atoms and nodes interchangeably.}, which in turn depends upon the local neighborhood around an atom. To capture these intricacies, we construct our state space using a set of semantic and topological node features.
    \begin{itemize}
        \item \textbf{Node type:} Each node $v \in \mathcal{V}$ is characterized by its type $\omega_v$. The type $\omega_v$ is a discrete variable and is useful in distinguishing particles of different characteristics within a system (Ex. two different types of atoms). We use \textit{one-hot encoding} to represent the node type. 
\item \textbf{Node potential energy:} Potential energy, being a scalar and extensive quantity, is additive in nature; that is, the potential energy of a system $U_\mathcal{G}$ is the sum of the potential energy of individual atoms. Consequently, the potential energy of a node can be a useful feature to identify the nodes that need to be displaced to reduce the overall energy.
        We denote the potential energy of node $v$ after $t$ steps as $U_v^t$ (Ex. see Fig.~\ref{fig:LJ_optim_Node_pe} for the distribution of potential energy per atom in an LJ system).
        \item \textbf{Neighborhood potential energy of a node:} As detailed earlier, the potential energy of an atom depends on its neighborhood (see Eq.~\ref{eq:LJ}). Thus, the energy of the neighborhood represents whether the atom is located in a relatively stable or unstable region. To this extent, we use the mean and the sum of the potential energy of atoms in the locality of the central atom as a node feature. We denote the sum of the potential energy of a node $v$'s neighborhood at step $t$ as $\textsc{Sum}(U^t_{\mathcal{N}_v})$, and the mean as $\textsc{Mean}(U^t_{\mathcal{N}_v})$. 
         
    \end{itemize}  
    Additionally, in order to capture the interactions of atoms, we use edge features. Specifically, we use the $L^1$ distance between two nodes $u$ and $v$ to characterize each edge $e_{uv}$. Finally, the empirical potentials modeling atomic structures present an \textit{equilibrium bond length} $|x^{equi}_{vu}|$ between two atoms; the distance at which these two atoms exhibit a minimum energy configuration. Note that $|x^{equi}_{vu}|$ for an atomic system can be directly obtained from the potential parameters (Ex. $2^{1/6}\beta$ for LJ; see Eq.~\ref{eq:LJ}). To represent this, we include an additional feature $|x^{equi}_{vu}| - |x_{vu}|$, where $|x_{vu}|$ is the bond length of the edge $e_{vu}$ connecting two atoms $v$ and $u$. This feature quantifies how much stretched/compressed the edge is from its equilibrium configuration.
Finally, the initial features of a node $v$ at step $t$ are: 
\begin{equation}
\label{eq:rep_nt}
    \mathbf{s}_v^t = \omega_v \mathbin\Vert U_v^t \mathbin\Vert \textsc{Sum}(U^t_{\mathcal{N}_v})\mathbin\Vert \textsc{Mean}(U^t_{\mathcal{N}_v})
\end{equation}
 where, $\mathbf{s}_v^t  \in  \mathbb{R}^{d_s}$ and $||$ denotes the \textit{concatenation} operation. Further, for an edge $e_{vu}$ with terminal nodes $v$ and $u$, its initial representation at step $t$ is:
\begin{equation}
\label{eq:rep_edge}
    \mathbf{s}_{e}^t = x_v-x_u \mathbin\Vert y_v-y_u\mathbin\Vert z_v-z_u \mathbin\Vert (|x^{equi}_{vu}| - |x_{vu}|)
\end{equation}
Using the above-designed node features, the state of a graph $\mathcal{G}$ at  step $t$ is denoted by a matrix $S_{\mathcal{G}^t} \in \mathbb{R}^{|\mathcal{V}|\times d_s}$ where each row $S_{\mathcal{G}^t}[i]=\mathbf{s}_i^t$. \\
 \textbf{Action:} We displace all the nodes of the graph differently at each step, hence the action space is continuous in our case and is represented as $\mathbf{a} \in \mathbb{R}^{|\mathcal{V}|\times d}$.\\
    \textbf{Reward:} Our objective is to reduce the overall potential energy of the system. One option is to define the reward $R^t$ at step $t\geq0$ as the reduction in potential energy of the system at step $t$, i.e., $U_{{\mathcal{G}}^{t}} - U_{{\mathcal{G}}^{t+1}}$. However, this definition of reward focuses on short-term improvements instead of long-term. In rough energy landscapes, the path to the global minima may involve crossing over several low-energy barriers. Hence, we use \textit{discounted rewards} $D^t$ to increase the probability of actions that lead to higher rewards in the long term. The discounted rewards are computed as the sum of the rewards over a \textit{trajectory} of actions with varying degrees of importance (short-term and long-term). Mathematically,
   \begin{equation}
        \label{Eq_Disc_Returns}
    D^t=R^t +\gamma R^{t+1}+\gamma^{2}R^{t+2}+\ldots= \sum_{k=0}^{T-t} \gamma^{k} R^{t+k}
    \vspace{-0.1in}
    \end{equation}
    where $T$ is the length of the trajectory and $\gamma \in (0,1]$ is a \textit{discounting factor} (hyper-parameter) describing how much we favor immediate rewards over the long-term future rewards.
    \looseness=-1\\    
\textbf{State transition:} 
 At each step $t$, all the nodes in the graph $\mathcal{G}^t$ are displaced based on the translation determined by the policy function $\pi$. The graph state thus transits from $S_{\mathcal{G}^t}$
 to $S_{{\mathcal{G}}^{t+1}}$. Since it is hard to model the transition dynamics $p(S_{{\mathcal{G}}^{t+1}} |S_{{\mathcal{G}}^{t}})$~\cite{hu2020graph}, we learn the policy in a \textit{model-free} approach. Sec.~\ref{sec:policyparam} discusses the details. 
\subsection{Neural method for policy representation}
\label{sec:policyparam}
\vspace{-0.05in}
The atoms in a system interact with other atoms in their neighborhood. In order to capture these interactions and infuse topological information, we parameterize our policy by a \gnn. At each step $t$, we first generate the representation of nodes using our proposed \gnn. These embeddings are next passed to an \textsc{Mlp} to generate a $|\mathcal{V}|\times d$-dimensional vector that represents the mean displacement for each node  in each direction. The entire network is then trained end-to-end. We now discuss each of these components in detail.

 \textbf{Graph neural network:}
Let $\mathbf{h}^0_v=\mathbf{s}_v^t$  denote the initial node representation of node $v$ and  $\mathbf{h}_{vu}^{0}$ denote the initial edge representation of edge $e_{vu}$. We perform $L$ layers of message passing to generate representations of nodes and edges. 
To generate the embedding for  node $v$ at layer $l+1$ we perform the following transformation:
\begin{equation}    
 \mathbf{h}_v^{l+1}=\sigma \left( {\textsc{Mlp}}\left( \mathbf{h}_v^{l} \mathbin\Vert \sum_{u \in \mathcal{N}_v} \mathbf{\mathbf{W}}^l_\mathcal{V} (\mathbf{h}_u^{l} \mathbin\Vert \mathbf{h}_{vu}^{l}) \right)\right)
\label{eq_hv}
\end{equation}
\vspace{-0.10in}

where $\mathbf{h}_v^{(l)}$ is the node embedding in layer $l$ and $\mathbf{h}_{vu}^{(l)}$ is the embedding of the edge between node $v$ and $u$ and $u \in \mathcal{N}_v$. $\mathbf{W}^l_\mathcal{V}$ is a trainable weight matrix and $\sigma$ is an activation function. The edge embedding  is computed as follows:
\vspace{-0.05in}
\begin{equation}
\mathbf{h}_{vu}^{l+1}=\sigma \left( {\textsc{Mlp}}\left( \mathbf{h}_{vu}^{l} \mathbin\Vert \mathbf{\mathbf{W}}^l_\mathcal{E}(\mathbf{h}_v^{l} \mathbin\Vert \mathbf{h}_{u}^{l}) \right)\right)
\end{equation}
where $\mathbf{h}^{(l)}_{vu}$ is edge embedding in layer $l$ for edge $e_{vu}$. $\mathbf{W}^l_\mathcal{E}$ is a trainable parameter.

Following $L$ layers of message passing, the final node representation of node $v$ in the $L^{th}$ layer is denoted by $\mathbf{h}_v^L \in \mathbb{R}^{d_h}$. Intuitively $\mathbf{h}_v^L$ characterizes $v$ using a combination of its own features and features aggregated from its  neighborhood. Note that the equations presented here correspond to the specific \gnn{} implementation used in \name{}{}. Indeed, we evaluate the effect of graph architecture by replacing our \gnn{} with other architectures such as graph attention network (GAT)~\cite{gat}, full graph network (FGN)~\cite{battaglia2018relational} later in Sec.~\ref{sec:graph_arch}.
\looseness=-1

As discussed, at each step $t$, the nodes in $\mathcal{G}^t$ are displaced based upon the action determined by policy function $\pi$.  Since our actions are continuous values, we must define the probability distribution over real-valued vectors. To this end, we employ \textit{multivariate Gaussian distribution}\footnote{Since we deal with $d$ dimensional action space, we use multivariate Gaussian.} $\mathcal{N}_d ( {\boldsymbol{\mu}, \boldsymbol{\Sigma} })$ for modeling the probability distribution over nodes. Here, $\boldsymbol{\mu} \in \mathbb{R}^d $ and $\mathbf{\Sigma} \in \mathbb{R}^{d\times d}$. Gaussian distribution is commonly used for continuous control in reinforcement learning~\cite{duan2016benchmarking,mnih2016asynchronous} since it is easy to sample from and its gradients can also be easily computed \cite{duan2016benchmarking, rumelhart1986learning}.

    For an action $\mathbf{a}_i \in \mathbb{R}^d $ on node $i$, we define the policy $\pi_\theta (\mathbf{a}_i|S_{\mathcal{G}^t})$ constructed from the distribution parameters $\boldsymbol{\mu}_i\in \mathbb{R}^d$ and $\boldsymbol{\Sigma}\in \mathbb{R}^{d\times d}$ as follows:

\begin{equation}
    \pi_\theta (\mathbf{a}_i|S_{\mathcal{G}^t}) =\left( \frac{1}{2 \pi} \right)^{d / 2} |{\bf \Sigma}|^{-1/2} \times  
    \mbox{exp} 
           \Bigg[ -\frac{1}{2} ({\mathbf{a}_i} - { \boldsymbol{\mu}_i})'{\bf \Sigma}^{-1} ({\mathbf{a}_i} - { \boldsymbol{\mu}_i}) \Bigg] 
\label{eq:decompose_prob}
\end{equation}

In the above equation, we parameterize mean $\boldsymbol{\mu}_i$ for node $i$ as: 
\begin{equation}
    \nonumber
\boldsymbol{\mu}_i = \mu_\theta(\mathbf{h}_{i}^{L})
\end{equation}

Recall  $\mathbf{h}_{i}^{L}$ is the embedding of node $i$ generated by \gnn{} in Eq.~\ref{eq_hv} and is a function of the state of the graph $\mathcal{G}^t$.
We do not parameterize $\boldsymbol{\Sigma}$ and instead use a fixed value, i.e., $\boldsymbol{\Sigma} = \alpha \times \mathbf{I} $ where $\alpha$ is a hyper-parameter and $\mathbf{I}\in \mathbb{R}^{d\times d}$ is identity matrix. This is done in order to simplify the learning process~\cite{turner2022adaptive}. Nonetheless, our design can be extended to output $\boldsymbol{\Sigma}$ as well. 
For a trajectory of length $T$, we sample actions for all nodes of the graph at each step $t$ using policy $\pi$. Consequently, for $\mathcal{G}^t$, we obtain an action vector $\mathbf{a}^t \in \mathbb{R}^{|\mathcal{V}|\times d }$. 

\subsection{Policy loss computation with baseline}
\label{sec:policy_loss_comp}
 Our goal is to learn parameters such that actions that lead to an overall reduction in energy are favored more over others.
  Towards this, we use \textit{REINFORCE gradient estimator} with baseline~\cite{reinforce} to optimize the parameters of our policy network. Specifically, we wish to maximize the reward obtained for the trajectory of length $T$ with discounted rewards $D^t$. To this end, we define a reward function $J(\pi_{\theta})$ as:
 \vspace{-0.1in}
\begin{equation}
     J(\pi_{\theta}) =\mathbb{E} \big[ \sum_{t=0}^T \left( D^t \right)  \big]
 \end{equation}
We, then, optimize $J(\pi_{\theta})$ with a \textit{baseline} $b$ as:
\vspace{-0.10in}
\begin{equation}
{
\nabla J(\pi_{\theta})= \left[ \sum_{t=0}^T \left( D^t -b(\mathcal{S}_{\mathcal{G}^t}) \right) \nabla_{\theta} log\pi_{\theta}(\mathbf{a}^t/\mathcal{S}_{\mathcal{G}^t})  \right]}
\label{eq:loss_rl}
\end{equation}

The role of a baseline $b(\mathcal{S}_{\mathcal{G}^t})$ is to estimate the difficulty of a state ${S}_{\mathcal{G}^t}$ (that is, how difficult it is to perform the task on ${S}_{\mathcal{G}^t}$ for the baseline) and better contextualize the rewards obtained by the actions generated by $\pi$~\cite{kool2018attention}. 
Empirically, it often reduces variance and speeds up learning. In our case, we use FIRE~\cite{bitzek2006structural} as the baseline since empirical performance obtained by FIRE was found to be better than other optimization techniques for rough landscapes (see Sec.~\ref{sec:hyperparameters}).
\vspace{-0.10in}
\subsection{Training and adaptation}
\vspace{-0.05in}

\textbf{Training phase:}
For a given set of training graphs, we optimize the parameters of the policy network $\pi_{\theta}$ for $T$ steps using  Eq.~\ref{eq:loss_rl}. \\
\textbf{Adaptation Phase:}
Once we obtain the trained model $\pi_{\theta}$, we adapt it to a target graph $\mathcal{G}_{target}$, which was unseen during training. Toward this, we optimize the parameters $\pi_{\theta}$ as well as the target graph $\mathcal{G}_{target}$ using Eq.~\ref{eq:loss_rl}. The central idea is to keep optimizing the graph structure for an extremely long trajectory (much larger than the training trajectory). However, training policy gradient with large values of $T$ can be difficult due to long-horizon problem~\cite{wang2020long}. To overcome this challenge, we sample a lower energy configuration (graph) obtained from the last three steps of the optimization trajectory (of length $T$) of the target graph $\mathcal{G}_{target}$. This sampled graph (configuration) now becomes the target graph, and we optimize this graph structure and the policy parameters. This process continues for a large number of steps($\gg T$). It enables the policy to adapt to a low-energy environment, completely unseen during the training, and successively get more stable configurations after each iteration without suffering from the long-horizon problem.
\looseness=-1
\vspace{-0.10in}
\section{Experiments}
\label{sec:experiments}
In this section, we evaluate the performance of \name{}{} to optimize atomic structures and compare it with other classical optimizers. We also analyze the effect of modifying the reward function, including additional features, and different graph architectures. Further, we show how the graph architecture enables generalization to unseen system sizes. 
\subsection{Experimental setup}
\textbf{$\bullet$ Simulation environment:} All the training and forward simulations are carried out in the JAX environment~\citep{schoenholz2020jax}. The graph architecture is implemented using the jraph package~\citep{jraph2020github}. \\
\textbf{Software packages:} numpy-1.24.1, jax-0.4.1, jax-md-0.2.24, jaxlib-0.4.1, jraph-0.0.6.dev0, flax-0.6.3, optax-0.1.4 \\
\textbf{Hardware:}
Processor: 2x E5-2680 v3 @2.5GHz/12-Core "Haswell" CPU
RAM: 62 GB"\\
\textbf{$\bullet$ Atomic systems and datasets:}
To evaluate the performance of \name, we consider three systems that are characterized by rough energy landscape, namely, (i) binary LJ mixture, (ii) Stillinger-Weber (SW) silicon, and (iii) calcium-silicate-hydrate (C-S-H) gel. The systems are discussed briefly below. The detailed equations of energy functions for these systems can be found in App.~\ref{sec:System_details}.\\
\textbf{Binary LJ:} We select a well-known binary mixture of two atom types with the atoms $A$ and $B$ in the ratio 80 and 20, respectively~\cite{kob1995testing}. The interactions in this system are pair-wise LJ (Eq.~\ref{eq:LJ}). However, this system is a good glass former and hence exhibits a large number of stable local minima. Further, the presence of two types of atoms makes optimization challenging for this system.\\
\textbf{SW Silicon (SW Si):} The empirical potential of SW Si is more complex, owing to the three-body angular term, thereby making the energy landscape more challenging to optimize~\citep{PhysRevB.31.5262}. Similar to the LJ system, SW Si also exhibits a large number of stable amorphous (disordered) states, although exhibiting a stable ordered crystalline state as well.\\
\textbf{Calcium silicate hydrate (C-S-H):} C-S-H is a coarse-grained model colloidal gel with interactions similar to LJ \citep{masoero2012nanostructure}, but of a higher degree polynomial. This structure is rarely found in an ordered state and, thus, similar to other systems, exhibits a rough landscape.\\ 
\textbf{Dataset generation:} The atomic structures corresponding to each of the systems are generated through molecular dynamics or Monte Carlo simulations at high temperatures. This ensures that the initial disordered structures are realistic and sampled from the high-energy regions of the landscape. For each system, $100$ atomic structures are selected randomly from the simulation. The detailed data generation procedure is given in App.~\ref{sec:System_details}.   \\      
\textbf{$\bullet$ Baselines:}
We compare the performance of \name{} with the following three classical optimizers, namely, (i) gradient descent~\cite{stillinger1986local}, (ii) Adam~\cite{kingma2014adam}, and (iii) FIRE~\cite{bitzek2006structural}. It is worth noting that while gradient descent and FIRE are widely used for atomic structures, Adam is rarely used. Nevertheless, due to the wide use of Adam for other optimization tasks, we include it in the present work. The hyper-parameters of the baseline have been chosen for each system to reach the lowest energy possible.\\
\textbf{$\bullet$ Evaluation metric:}
Since the goal of the present work is to find the most stable structure starting from a random initial structure, we use the potential energy of the structure as the metric to evaluate the performance of the algorithms. A more stable structure corresponds to lower energies, with the global minima exhibiting the lowest energy structure. Note that the energy for each of the systems considered is computed using the respective empirical potential. Additionally, to evaluate the performance of the model during the training phase, we compute the change in energy during a given trajectory of length $T$ on the validation graphs. Specifically, at different training epochs, we calculate the average reduction in energy of the system in $20$ optimization steps (5 steps longer than the training trajectory), $<E_{20}-E_0>$, where $E_{20}$ is the energy at the $20^{th}$ step and $E_0$ is the energy of the initial configuration from the validation set. \\
\textbf{$\bullet$ Model architecture and training setup:}
All the hyperparameters of the model are given in Tabs.~\ref{tab:hyperparameters} and ~\ref{tab:Baseline hyperparamenter} in App.~\ref{sec:hyperparameters}. For the \gnn{}, the node and edge embeddings are chosen to be of size 48 with a single message passing layer. All MLPs, except the initial node embedding generation MLP and the final displacement prediction MLP, have two hidden layers, each having $48$ hidden layer units. The initial node embedding generation MLP has an additional batch-normalization layer, while the final MLP has four hidden layers. Leaky-ReLU is used for all the MLPs as the activation function.

For each system, a dataset of $100$ initial states of the environment sampled from the simulation, randomly split into $75:25$ training and validation sets, respectively, are used to train the model. During training, at each epoch, a trajectory length of $T=15$ is used to compute the reward function $J(\pi_{\theta})$, and the batch-average loss is used to compute the policy gradient. Validation is performed for the trained model on a trajectory of $T=20$ steps by selecting graphs randomly from the validation set. Note that validation is performed every $20$ epochs. For the adaptation of the trained model to obtain minimum energy, $10$ new \textit{target structures(graphs)}, that were not part of the training or validation sets and randomly sampled from the simulation, were used as starting structures. Adaptations of these graphs were carried out for $1000$ epochs, with each epoch having a trajectory length of 15 steps. Further, for each structure, the adaptation of \name{}{} was performed on $10$ random seeds, and the model that gave the minimum energy structure was selected. For each system, the mean of the minimum energy obtained on the 10 structures and the lowest minima among the 10 structures are reported.

For the baselines, the minimization was carried out for $1000$ steps in the case of LJ and SW Si, and for $2000$ steps in the case of C-S-H. In all the cases, the steps were long enough to ensure that the energy of the structures obtained by baselines was saturated. Similar to \name{}{}, the minimization was performed on the same $10$ configurations, and both the mean minimum energy and lowest minimum energy obtained are reported.
\looseness=-1

\subsection{\name{}: Comparison with baselines}
\label{sec:comparison}
First, we analyze the performance of \name{} on the three systems, namely, LJ, C-S-H, and SW Silicon, to optimize the structures. Figs.~\ref{fig:Val_curves},~\ref{fig:pre_train_reward} in Appendix show the validation and reward curves, respectively, for these models during the training. Table~\ref{tab:Energies} shows the minimum and mean energies obtained by \name{} compared to the baselines for the three systems on 10 initial structures. We note that \name{} achieves better minima than the baselines for LJ, C-S-H, and SW Silicon systems, both in terms of the minimum energy achieved and the mean over 10 structures. We also note that both FIRE and Adam consistently outperform gradient descent. Interestingly, Adam  outperforms FIRE on SW Silicon. For the C-S-H system, Adam and FIRE exhibit comparable performance, while for the LJ system, FIRE outperforms Adam. Nevertheless, we observe that \name{} exhibits notably better performance than all the other classical optimization algorithms in obtaining a stable low-energy structure. The superior performance of \name{} could be attributed to several components, such as discounted rewards and graph topology.  While discounted reward allows us to overcome local barriers, graph-based modeling enables richer characterization of atomistic configurations through topology.

\looseness=-1
\begin{table*}[ht]
\centering
\scalebox{1}{
    \begin{tabular}{|c|c|c|c|c|c|}
         \toprule
                 \textbf{Atomic system} & \textbf{Metric} & \textbf{Gradient Descent} & \textbf{FIRE} & \textbf{Adam} & \textbf{\name}\\
         \midrule
         \multirow{2}{*}{LJ ($\varepsilon$ \ units) }& Min &-799.53 &-813.66
         & -808.62 & \textbf{-815.63} \\\cline{2-6}
          & Mean & -795.38 & -806.29 & -801.96 & \textbf{-811.99}\\
         \hline
         \multirow{2}{*}{C-S-H (kcal/mol)}  & Min & -1583539.3 & -1637194.1 &-1622905.9 & \textbf{-1671916.8} \\\cline{2-6}
         & Mean & -1548798.6 & -1588792.4 & -1596680.4 & \textbf{-1648965.9} \\
        \hline
         \multirow{2}{*}{SW Silicon (eV)} & Min & -249.22 &-256.98 &-258.86 &\textbf{-259.94}\\\cline{2-6}
         & Mean & -247.56 & -256.37 & -256.93 & \textbf{-257.35} \\
        \hline
        \end{tabular}}
        \caption{Comparison of \name{} with classical optimization algorithms for LJ, C-S-H, and SW Silicon systems. For each system, the minimum and mean energies are evaluated on 10 random initial structures.}
        \label{tab:Energies}
       \end{table*}

\vspace{-0.10in}
\subsection{Effect of baseline and additional components}
\label{sec:additional_features}
\vspace{-0.05in}
\begin{figure*}[!ht]
    \centering
    \includegraphics[width=\textwidth]{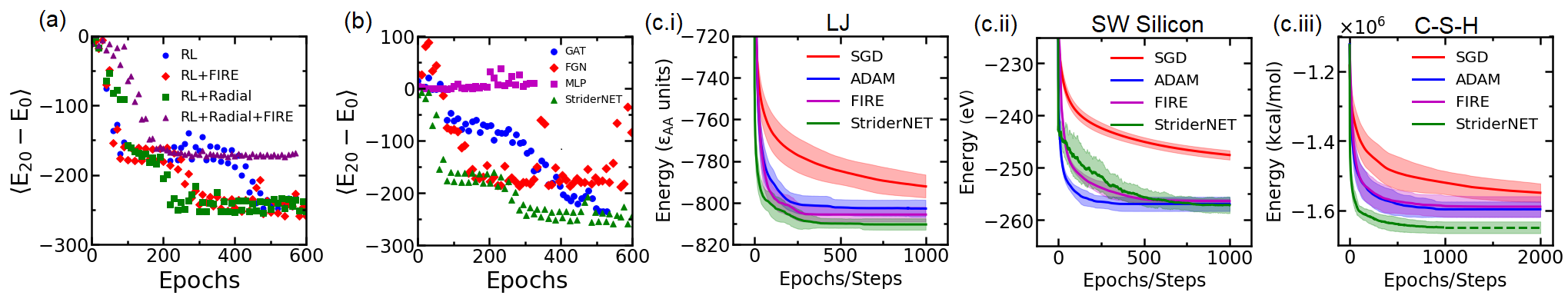}
    \caption{(a) Validation curve during training for different models. (b) Comparison of different graph architectures for RL algorithm, namely, GAT, FGN, \name{}, and MLP. (c) Evolution of energy during adaptation of \name{} for: (i) LJ, (ii) SW Silicon, and (iii) C-S-H system, respectively. The curve represents the mean over 10 structures, and the shaded regions represent the standard deviation. Note that the \name{}{} for C-S-H is run only for 1000 epochs and the dotted line represents the value at the $1000^{th}$ step.}
    \label{fig:Combined_models_archs_opt_curves}
    \end{figure*}

Now, we analyze the role of several components in \name{}{} such as the use of FIRE as baseline in eq.~\ref{eq:loss_rl} and additional features towards its performance. \name{}{} uses FIRE as  baseline during training and adaptation. To analyze the effect of baseline, the first variation, termed RL, discards the FIRE baseline and is trained with $b(\mathcal{S}_{\mathcal{G}^t}){=}0$. The second variation, termed RL+FIRE, equivalent to the \name, uses FIRE as a baseline during the training. The third variation, termed RL+Radial, employs vanilla RL with the radial symmetry functions~\cite{behler2011atom} as an additional node input feature for the \gnn{s}, which has been shown to provide excellent neighborhood representation for atomic structures. The final variation, termed RL+Radial+FIRE, uses both FIRE as the baseline and the radial functions as additional input features for the nodes in the \gnn{s} for better neighborhood representation.

Fig.~\ref{fig:Combined_models_archs_opt_curves}(a) in appendix shows the validation curve of the trained models with the above-mentioned variations. We observe that the best performance is achieved by RL+FIRE and RL+Radial. Note that including radial features (RL+Radial) makes the computation more expensive for this model~\cite{behler2011atom}. We also observe that RL performs similarly to RL+FIRE, although for larger epochs. However, the forward trajectory of the RL without baseline occasionally exhibits instability, whereas the RL+FIRE exhibits highly stable inference. We observe that RL+Radial+FIRE shows poorer performance than RL+FIRE and RL+Radial. Altogether, we observe that the \name, represented by RL+FIRE, represents the optimal model in terms of computational efficiency and inference.
\looseness=-1
\vspace{-0.10in}
\subsection{Graph Architectures: MLP, GAT, FGN, \name}
\vspace{-0.05in}
\label{sec:graph_arch}
We evaluate the role of the \gnn{s} architecture on the performance of \name. To this extent, we compare three models  with different graph architectures, namely, GAT, FGN, and \name, which has our own architecture (see Sec.~\ref{sec:policyparam}). In order to evaluate the role of \gnn{s}, we also trained a model with a fully-connected feed-forward multilayer perceptron (MLP). In Fig.~\ref{fig:Combined_models_archs_opt_curves}(b) we observe that the proposed \gnn{} architecture in \name{}{} provides superior performance, although GAT also leads to similar performance for larger epochs. We note that the FGN architecture is unable to achieve comparable performance. Interestingly, the MLP-based model fails to train and shows no reduction in energy, even at large epochs. This suggests that the topology and neighborhood information, as captured by the \gnn{} through message passing plays a crucial role in the performance of \name.
\subsection{Model adaptation}
\vspace{-0.05in}
\label{sec:adaptation}
Now, we analyze the evolution of the energy of a structure during adaptation. Fig.~\ref{fig:Combined_models_archs_opt_curves}(c) shows the performance of \name{}{} along with the baselines on $10$ structures. It should be noted that for \name{}{}, the adaptation of the trained model involves back-propagation; hence, the evolution of energy is plotted with the number of epochs in this case. In the case of both LJ and C-S-H systems, we observe that \name{}{} consistently exhibits lower energy than other models. In the case of SW Si, we observe that \name{}{}, although initially exhibiting higher energy, eventually outperforms other models. Thus, we observe that the model adaptation on an unseen target graph structure allows \name{}{} to outperform classical optimization algorithms.
\looseness=-1
\vspace{-0.10in}
\subsection{Inductivity to varying system sizes}
\vspace{-0.05in}
\label{sec:inductivity}

\begin{table}
\centering
\scalebox{1}{
\begin{tabular}{|c|c|c|c|c|c|}\hline 
  \textbf{Number of atoms} & \textbf{Metric} & \textbf{Gradient descent} & \textbf{Adam} & \textbf{FIRE} & \textbf{\name} \\  \hline

  \multirow{2}{*}{25} & Min.    & -6.94 & -7.00 & -6.99 & \textbf{-7.08}  \\ \cline{2-6}
     & Mean    &  -6.79 & -6.91 & -6.81 & \textbf{-6.97}\\ \hline
  
  \multirow{2}{*}{50} & Min.    & -7.67 & -7.70 & -7.67 & \textbf{-7.77}  \\ \cline{2-6}
    & Mean    &  -7.57 & -7.62 & -7.63 & \textbf{-7.71}\\ \hline
  
  \multirow{2}{*}{100} & Min.    & -8.00 & -8.09 & -8.14 & \textbf{-8.16}  \\ \cline{2-6}
    & Mean    &  -7.92 & -8.03 & -8.06 & \textbf{-8.12}\\ \hline

  \multirow{2}{*}{250} & Min.    & -8.02 & \textbf{-8.15} & \textbf{-8.15} & \textbf{-8.15}  \\ \cline{2-6}
    & Mean    &  -7.98 & -8.10 & -8.11 & \textbf{-8.13}\\ \hline

  \multirow{2}{*}{500} & Min.    & -8.02 & -8.14 & -8.14 & \textbf{-8.16}  \\ \cline{2-6}
    & Mean    &  -7.99 & -8.12 & -8.12 & \textbf{-8.14}\\ \hline

  \multirow{2}{*}{1000} & Min.    & -8.00 & -8.13 & \textbf{-8.14} & -8.13  \\ \cline{2-6}
     & Mean    &  -7.98 & \textbf{-8.12} & \textbf{-8.12} & \textbf{-8.12}\\ \hline

\end{tabular}}
 \vspace{0.10in}
\caption{Minimum energy obtained by adaptation of \name{}{} trained on a $100$-atom LJ system to varying system sizes. For comparison among multiple sizes, total energy normalized by the number of atoms in the system is shown.}
\label{tab:inductivity}
\end{table}

Finally, we evaluate the ability of \name{}{} trained on a given graph size to adapt to unseen graph sizes. To this extent, we consider the \name{}{} trained for the LJ system having 100 atoms and adapt it to different system sizes with $N={25,50,250,500,1000}$. Table~\ref{tab:inductivity} shows the performance of \name{}{} on all the system sizes. Interestingly, for all structures from $25$ to $500$ atoms, we observe that \name{}{} gives the best performance in terms of both the overall minimum and the mean of the minimum energies of 10 structures. For the 1000 atom system, we observe that \name{}{} gives the same performance as Adam and FIRE for mean energy, while FIRE outperforms Adam and \name{}{} in terms of the minimum energy achieved. However, it is worth noting that \name{}{} gives comparable performance for the mean energy even for $1000$ atom structures; that is one order larger than the trained graph.
\looseness=-1
\vspace{-0.10in}
\section{Conclusion}
\label{sec:conc}
\vspace{-0.05in}
In this work, we present \name, a graph reinforcement learning approach that enables the optimization of atomic structures on a rough landscape. We evaluate the model on three systems, namely, LJ, C-S-H, and SW Silicon, and show that \name{}{} outperforms the classical optimization algorithms such as gradient descent, FIRE, and Adam. We also show that the model exhibits inductivity to completely unseen system sizes; \name{}{} trained on 100 atom yields superior performance for a 500 atom system. Altogether, \name{}{} presents a promising framework to optimize atomic structures.
\looseness=-1

\textbf{Limitations and future work:} Although promising, \name{}{} is limited to a relatively small number of atoms. Scaling it to a larger number of atoms presents a major computational challenge. Further, although \name{}{} outperformed classical local optimizers, the energy reached by \name{}{} is not the global minimum. Thus, there is further scope for improvement that enables one to discover the global minimum in these structures.
\looseness=-1

\bibliographystyle{abbrvnat}
\bibliography{biblio}

\clearpage

\section{Notations}
\label{sec:notation}
All the notations used in this work are outlined in Tab.~\ref{tab:notation}.

\begin{table}[!ht]
    \centering
    \scalebox{0.9}{
        \begin{tabular}{m{0.25\linewidth}  m{0.7\linewidth}}
         \toprule
         \textbf{Symbol} & {\textbf{Meaning}} \\
         \midrule
     
         $\mathcal{G}^t$ & Graph at step $t$\\ \cmidrule(lr){1-2}
         $\mathcal{V}$& Node set \\ \cmidrule(lr){1-2}
         $\mathcal{E}^t$& Edge set at step $t$\\ \cmidrule(lr){1-2}
         $\mathcal{S}_{{\mathcal{G}}^t}$& State of Graph at step $t$\\ \cmidrule(lr){1-2}
         $\mathcal{N}_v$&Neighboring nodes of node $v$ \\ \cmidrule(lr){1-2}
         $U_v$ & Potential energy of node $v$\\\cmidrule(lr){1-2} 
         $U^t_v$ & Potential energy of node $v$ at step $t$ for graph $\mathcal{G}^t$\\\cmidrule(lr){1-2}
         $U_{\mathcal{G}^t}$ & Potential energy of graph $\mathcal{G}$ at step $t$\\ \cmidrule(lr){1-2}
         $e$ & Edge $e \in \mathcal{E}$ \\ \cmidrule(lr){1-2}
         $d$ & Number of  Dimensions in the system  \\\cmidrule(lr){1-2}
         $\mathbf{s}^t_v$ & Initial feature representation of node $v$ at step $t$  \\ \cmidrule(lr){1-2}
         $T$ & Length of trajectory \\ \cmidrule(lr){1-2}        
         $\pi$ & Policy function\\ \cmidrule(lr){1-2}
         $\mathbf{a}$ & Action vector for all nodes of a graph. $\mathbf{a} \in \mathbb{R}^{|\mathcal{V}| \times d}$ \\ \cmidrule(lr){1-2} 
         ${\mu}_i $ & Predicted mean displacement for the $i^{th}$ node. ${\mu}_i \in \mathbb{R}^d$  \\ \cmidrule(lr){1-2}
         $\mathbf{\Sigma} $ & Covariance Matrix. $\mathbf{\Sigma} \in \mathbb{R}^{d\times d}$ \\ \cmidrule(lr){1-2}. 
     \end{tabular}
        }
            \caption{Notations used in the paper}
    \label{tab:notation}
\end{table}

\section{System Details}
\label{sec:System_details}
\subsection{Binary Lennard-Jones (LJ)}

The system has two types of particles with composition $A_{80}B_{20}$ consisting of total N(=25,50,100,250,500) particles in a cubic ensemble with periodic boundaries. The interaction between the particles is governed by
\begin{equation}
\label{eq:lj}
V_{\mathrm{LJ}}(r)=4 \varepsilon\left[\left(\frac{\sigma}{r}\right)^{12}-\left(\frac{\sigma}{r}\right)^{6}\right]
\end{equation}
where $r$ refers to the distance between two particles, $\sigma$ is the distance at which inter-particle potential energy is minimum and $\varepsilon$ refers to the depth of the potential well. Here, we use the LJ parameters $\varepsilon_{AA}=1.0$, $\varepsilon_{AB} = 1.5$, $\varepsilon_{BB}= 0.5$, $\sigma_{AA} = 1.0$, $\sigma_{AB}= 0.8$ and $\sigma_{BB}= 0.88$. The mass for all particles is set to $1.0$. All the quantities are expressed in reduced units with respect to $\sigma_{AA}$,  $\varepsilon_{AA}$, and Boltzmann constant $k_B$. We set the interaction cutoff $r_c=2.5\sigma$~\cite{singh2013ultrastable} and the time step $dt=0.003$ for simulations. 

We perform all the molecular dynamic simulations at constant volume and temperature. For preparing the initial high energy  structures, the ensemble is taken to a high temperature $T=2.0$ where it equilibrates in the liquid state. Once it equilibrates, 100 random configurations are sampled.  

\subsection{Stillinger Weber (SW) Silicon}
The system consists of N=64 particles in a cubic ensemble with periodic boundaries interacting via the Stillinger Weber(SW) potential, as given by the following equation.

\begin{align}
E & =\sum_i \sum_{j>i} \phi_2\left(r_{i j}\right)+\sum_i \sum_{j \neq i} \sum_{k>j} \phi_3\left(r_{i j}, r_{i k}, \theta_{i j k}\right) \nonumber\\
\phi_2\left(r_{i j}\right) & =A_{i j} \epsilon_{i j}\left[B_{i j}\left(\frac{\sigma_{i j}}{r_{i j}}\right)^{p_{i j}}-\left(\frac{\sigma_{i j}}{r_{i j}}\right)^{q_{i j}}\right] \exp \left(\frac{\sigma_{i j}}{r_{i j}-a_{i j} \sigma_{i j}}\right) \\
\end{align}

\begin{equation}
\phi_3\left(r_{i j}, r_{i k}, \theta_{i j k}\right) =\lambda_{i j k} \epsilon_{i j k}\left[\cos \theta_{i j k}-\cos \theta_{0 i j k}\right]^2 \exp \left(\frac{\gamma_{i j} \sigma_{i j}}{r_{i j}-a_{i j} \sigma_{i j}}\right) \times \\ \exp \left(\frac{\gamma_{i k} \sigma_{i k}}{r_{i k}-a_{i k} \sigma_{i k}}\right)\nonumber
\end{equation}

where $\phi _{2}$ is the two body term and $\phi _{3}$ is the three-body angle term. The following are the standard parameters\citep{stillinger1986local} used in the equation:
\begin{table}[!ht]
    \centering
    \scalebox{0.845}{
    \begin{tabular}{|l|l|l|l|l|l|l|l|l|l|l|}
         \hline
         Parameter&$\varepsilon$&$\sigma$& A & B & p & q & a & $\lambda$ & $\gamma$&$cos \theta _{0}$  \\
         \hline
         Value&2.1683 eV&2.0951 \r{A} &7.0495&0.6022&4&0&1.80&21.0&1.20&-1/3\\
         \hline
    \end{tabular}}
    \caption{Parameters for Stillinger weber potential }
    \label{tab:SW_params}
\end{table}

\par
We equilibrate the system at a high temperature of T=3500 K in an isochoric-isothermal (NVT) ensemble to obtain the initial high-energy configurations.

\subsection{Calcium silicate hydrate (C-S-H) gel}

Calcium silicate hydrate(C-S-H) is the binding phase in concrete. C-S-H is known to govern various properties of concrete, including strength and creep. The coarse-grained colloidal gel model of C-S-H used in this work was proposed by Masoero et al.\cite{masoero2012nanostructure}. The model has been studied extensively and found to be capable of simulating the realistic mesoscale structure of C-S-H as well as long-term creep behavior\cite{liu2019long,liu2021predicting}.

The C-S-H particles interact with each other via a generalized Lennard-Jones interaction potential as given by the following equation:
\begin{equation}
    \label{Eqn_CSH_potential}
    U_{ij}(r_{ij})=4\varepsilon\Bigg[\left(\frac{\sigma}{r_{ij}} \right)^{2\alpha} - \left(\frac{\sigma}{r_{ij}}\right)^\alpha \Bigg]
\end{equation}

Where $U_{ij}$ is the interaction potential energy between any pair particles 'i' and 'j', $r_{ij}$ is the distance between the particles, and $\sigma$ is the grain diameter which is taken to be 5 nm in the model. $\alpha$ is a parameter that controls the potential well's narrowness. $\alpha$ is chosen to be 14 such that the tensile strain at failure is close to that obtained in previous simulations of bulk C--S--H. $\varepsilon$ is the potential well's energy depth. The energy depth is given by $\varepsilon=A_0\sigma^3$, where $A_0=kE$ and E is the young's modulus of bulk C--S--H grain, which is around 63.6 GPa~\cite{manzano2013shear} and k=0.0023324.

\subsubsection{Preparation of C-S-H by GCMC simulations and obtaining high energy states}
During the hydration process, the chemical reaction between the cement and electrolytes in water occurs via a dissolution-precipitation reaction. The grand canonical Monte Carlo (GCMC) simulations mimic the precipitation process during the hydration of cement. The C-S-H particles are iteratively inserted in an empty cubic box ensemble with periodic boundary conditions. In each step of the simulation, `X' attempts of grain exchanges(i.e., insertions and deletions) are performed, which is followed by `M' attempts of randomly displacing the grains to achieve a more stable configuration. The following equation gives the Monte Carlo acceptance probability according to the Metropolis algorithm:
\begin{equation}
    \label{Eqn_GCMC_P_Acceptance}
    P_{acceptance}=min\Bigg\{1,exp\bigg[-\Bigg(\Delta U-\frac{\mu\lambda}{k_BT}\Bigg)\Bigg]\Bigg\} 
\end{equation}

where $\Delta U$ is the change in energy after the Monte Carlo trial move, $\mu$ is the chemical potential which represents the free energy gained by the formation of C-S-H hydrates, $\lambda$ is the variation in the number of C-S-H particles, $k_B$ is Boltzmann constant. $T$ is the temperature of an infinite reservoir source. The chemical potential of the reservoir is kept as $2k_BT$ as per the previous studies\cite{ioannidou2016crucial,liu2019effects}. The GCMC steps are performed until the no. of inserted C-S-H grains reaches saturation. The simulations are performed at a temperature of T=300 K. The final saturated configurations so obtained are relaxed in the isothermal-isobaric (NPT) ensemble at 300 K and zero pressure for 50 ns to release ant macroscopic tensile stress induced during GCMC simulation. Finally, energy minimization is performed to reach the inherent state of the configuration.
\par
Next, the obtained structure is taken to a high temperature of T=1000K in an isothermal-isochoric (NVT) ensemble and allowed to equilibrate. Once it equilibrates, 100 random configurations are sampled. 
The GCMC simulation was performed in  Large-scale Atomic/Molecular Massively Parallel Simulator (LAMMPS)~\cite{LAMMPS} software.

\section{Reward and validation curves of StriderNet}
\begin{figure}[!ht]
    \centering
    \includegraphics[width=\columnwidth]{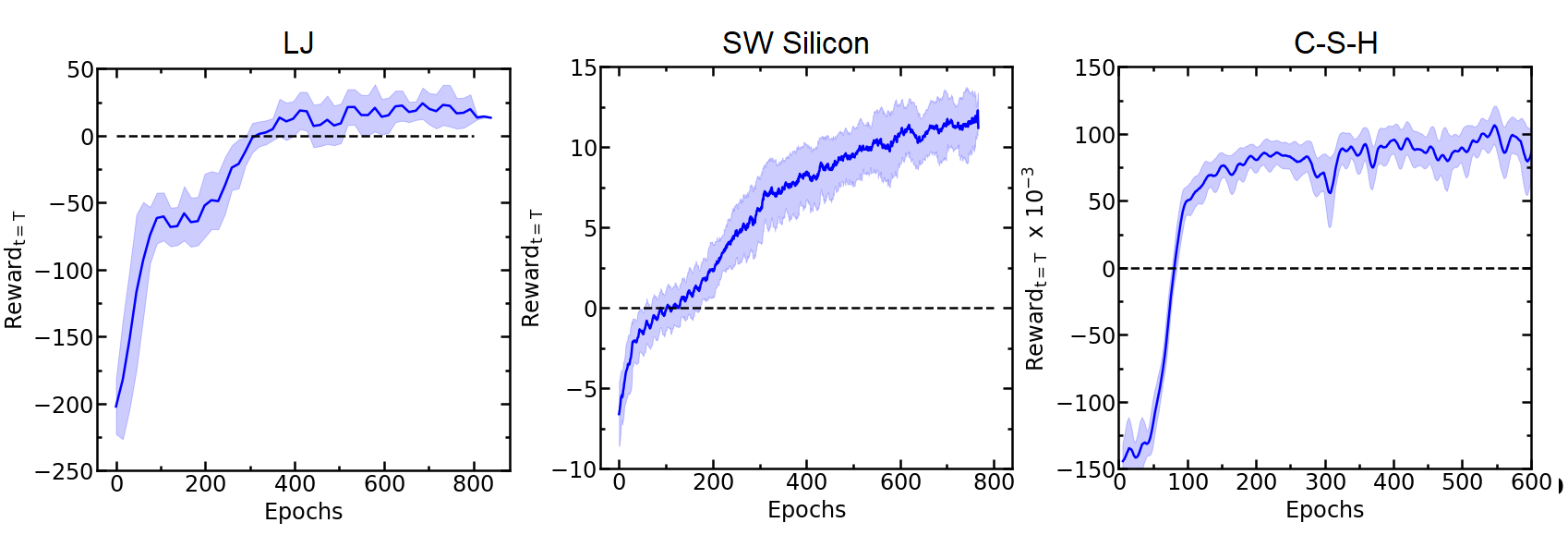}
    \caption{Reward at the end of trajectory during the training of \name{} for LJ, SW Si, C-S-H systems.}
    \label{fig:pre_train_reward}
\end{figure}

\begin{figure}[!ht]
    \centering
    \includegraphics[width=\textwidth]{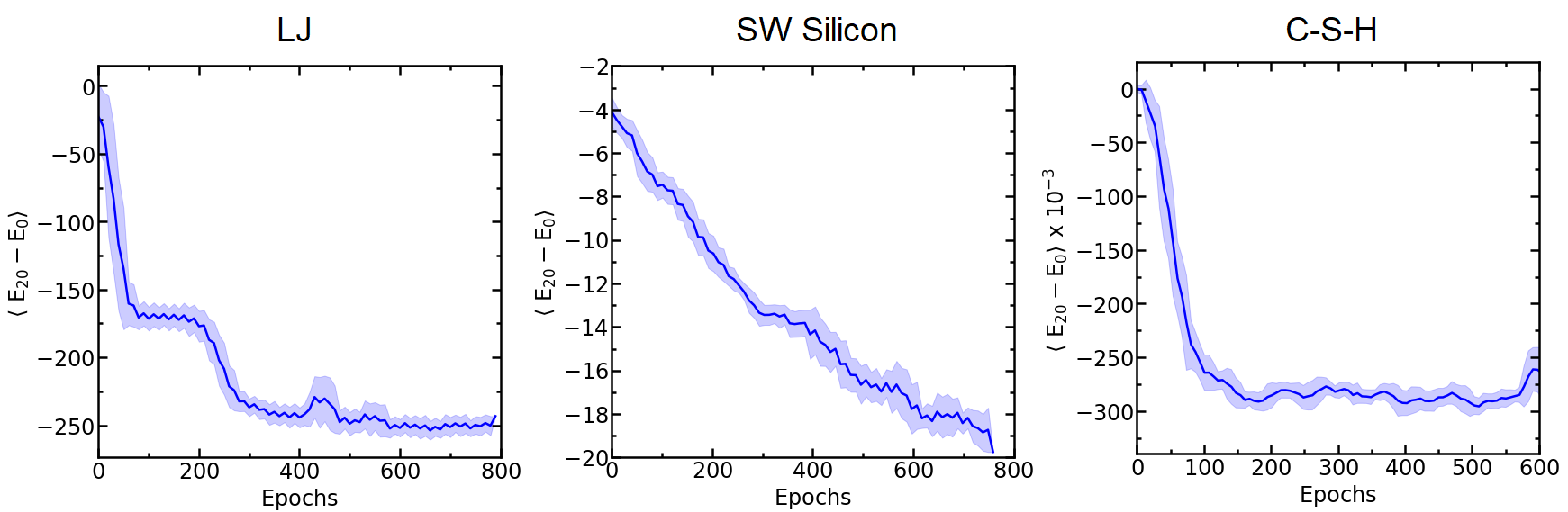}
    \vspace{-0.15in}
    \caption{Validation curves: Average reduction in energy in 20 steps of optimization during the training of \name{} for LJ, SW Si, and C-S-H.}
    \label{fig:Val_curves}
    \vspace{-0.20in}
\end{figure}

Figure~\ref{fig:pre_train_reward} shows the reward at the end of each of the validation trajectories for \name{}{} trained on LJ, SW Si, and C-S-H systems. Positive values of the rewards suggest that the model has outperformed FIRE on the validation graphs. Figure~\ref{fig:Val_curves} shows the difference between the energy at the beginning and the end of the trajectory on the validation set. We observe that the curve saturates for both LJ and C-S-H systems. However, SW Si exhibits a further downward trend after 800 epochs. It is worth noting that the SW Si has a tendency for crystallization and exhibits a global minimum crystalline structure. Thus, it would be worth exploring further on continuing the training of the SW Si systems towards exploration of a lower minimum.

\section{Hyperparameters of \name{}{} and baselines}
\label{sec:hyperparameters}
Hyperparameters of \name{}{} are included in Tab.~\ref{tab:hyperparameters}. Further, the hyperparameters associated with the baselines, namely, FIRE, Adam, and gradient descent are included in Tab.~\ref{tab:Baseline hyperparamenter}. To reduce computational overhead, we run baseline only on the initial state and use that value across all steps in the trajectory during the training of \name{}{}.
\begin{table}[!ht]
    \centering
    \scalebox{0.9}{\begin{tabular}{ |l|l|  }
         \hline
         \multicolumn{2}{|c|}{Hyper-parameters} \\
         \hline
         PARAMETER& VALUE \\
         \hline
          Edge embedding size
          &   48\\
         \hline
          Node embedding size
          &   48\\
         \hline
          Initial node embedding MLP $f_a$ layers
          &   3\\
         \hline
          Initial edge embedding MLP $f_b$ layers
          &   2\\
         \hline
          Edge update MLP layers
          &   2\\
         \hline
          Node update MLP layers
          &   2\\
         \hline
          Node displacement MLP layers
          &   4\\
         \hline
          Message passing steps($L$)
          &   1\\
         \hline
         Batch-norm layer decay rate for the exponential moving average
          & 0.9 \\
         \hline
         Trajectory length($T$) 
          &   15\\
         \hline
         Gradient accumulation steps
          &   2\\
         \hline
          Graphs training batch size
          &   4\\
         \hline
          Edge to node aggregation function
          &   Mean\\
         \hline
          Activation functions(all MLPs)
          &   Leaky ReLU\\
         \hline
          Multivariate gaussian constant factor($\alpha$)
          &   $10^{-5}$\\
         \hline
          Rewards discount factor ($\gamma$)
          &   0.9\\
         \hline
         Training optimizer
          &   Adam\\
         \hline
          Training optimizer learning rate
          &   0.005\\
         \hline
          Node displacement MLP neighborhood aggregation
          &   Mean\\
          \hline
          Predicted displacement scaling factor
          &   2.0 (LJ), 2.0 (SW Si),5.0 (C-S-H)\\
          \hline
          Gradient clipping
          &   0.1\\
          \hline
    \end{tabular}}
    \caption{Hyper-parameters of \name{}{}}
    \label{tab:hyperparameters}
\end{table}

\begin{table}[!ht]
    \centering
    \begin{tabular}{|l|l|l|l|l|}
        \hline
        Baseline&Parameter&LJ&SW SIlicon&CSH  \\ \hline
        Gradient descent&Learning rate&$5\times10^{-4}$ &$10^{-3}$ &$5 \times 10^{-4}$   \\ \hline
        
        \multirow{5}{*}{Adam}&Learning rate& 0.05&0.1 &1.0\\ \cline{2-5}
        &$\beta_{1}$& \multicolumn{3}{|c|}{0.9}\\ \cline{2-5}
        &$\beta_{2}$& \multicolumn{3}{|c|}{0.999}\\ \cline{2-5}
        &$\varepsilon$& \multicolumn{3}{|c|}{$10^{-8}$}\\ \cline{2-5}
        &$\overline{\varepsilon}  $ & \multicolumn{3}{|c|}{0.0}\\ \hline
         \multirow{7}{*}{FIRE}&$dt_{start}$&0.01 &0.5&$5\times10^{-3}$\\  \cline{2-5}
        &$dt_{max}$& \multicolumn{3}{|c|}{0.4}\\  \cline{2-5}
        &$N_{min} $& \multicolumn{3}{|c|}{5}\\  \cline{2-5}
        &$f_{\alpha}$& \multicolumn{3}{|c|}{0.99}\\  \cline{2-5}
        &$f_{dec}  $ & \multicolumn{3}{|c|}
        {0.5}\\  \cline{2-5}
        &$f_{inc}  $ & \multicolumn{3}{|c|}
        {1.1}\\  \cline{2-5}
        
        &$\alpha_{start}  $ & \multicolumn{3}{|c|}{0.1}\\ \hline

    \end{tabular}
    \caption{Baselines hyperparameters}
    \label{tab:Baseline hyperparamenter}
\end{table}

\end{document}